\documentclass{tlp}

\usepackage{amsmath}
\usepackage{graphicx}
\usepackage{multirow}
\usepackage{algorithm}
\usepackage{algorithmic}
\usepackage{booktabs}
\usepackage{multirow}
\begin{document}

\lefttitle{A Neurosymbolic Framework for Bias Correction in CNNs}

\jnlPage{TO DO}{TO DO}
\jnlDoiYr{2024}
\doival{TO DO}

\title[A Neurosymbolic Framework for Bias Correction in CNNs]{A Neurosymbolic Framework for Bias Correction in Convolutional Neural Networks\thanks{This work was supported by US NSF Grant IIS 1910131, US DoD, industry grants, EPSRC grant EP/T026952/1, \emph{AISEC: AI Secure and Explainable by Construction} and the EPSRC DTP Scholarship for N.~Ślusarz. We thank the members of the ALPS lab at UT Dallas for insightful discussions. We are grateful to the anonymous reviewers for their insightful comments and suggestions for improvement. Competing interests: The author(s) declare none}}

\begin{authgrp}
\author{\sn{Parth Padalkar}}
\affiliation{The University of Texas at Dallas}
\author{\sn{Natalia Ślusarz}}
\affiliation{Heriot-Watt University}
\author{\sn{Ekaterina Komendantskaya}}
\affiliation{Southampton University, Heriot-Watt University}
\author{\sn{Gopal Gupta}}
\affiliation{The University of Texas at Dallas}
\end{authgrp}

\history{\sub{13 05 2024;} \rev{18 07 2024} 
%\acc{; xx xx xxxx}
}

\maketitle

\begin{abstract}

Recent efforts in interpreting Convolutional Neural Networks (CNNs) focus on translating the activation of CNN filters into a stratified Answer Set Program (ASP) rule-sets. The CNN filters are known to capture high-level image concepts, thus the predicates in the rule-set are mapped to the concept that their corresponding filter represents. Hence, the rule-set exemplifies the decision-making process of the CNN w.r.t the concepts that it learns for any image classification task. These rule-sets help understand the biases in CNNs, although correcting the biases remains a challenge. We introduce a neurosymbolic framework called NeSyBiCor for bias correction in a trained CNN. Given symbolic concepts, as ASP constraints, that the CNN is biased towards, we convert the concepts to their corresponding vector representations. Then, the CNN is retrained using our novel semantic similarity loss that pushes the filters away from (or towards) learning the desired/undesired concepts. The final ASP rule-set obtained after retraining, satisfies the constraints to a high degree, thus showing the revision in the knowledge of the CNN. We demonstrate that our NeSyBiCor framework successfully corrects the biases of CNNs trained with subsets of classes from the \textit{Places} dataset while sacrificing minimal accuracy and improving interpretability.
% by greatly decreasing the size of the final bias-corrected rule-set w.r.t. the initial rule-set.
\end{abstract}


\section{Introduction}
Deep learning models, particularly those designed for vision-based tasks such as image classification, have proliferated across various domains, demonstrating remarkable performance in discerning intricate patterns in visual data. However, this widespread adoption comes with a caveat---these models are inherently susceptible to biases present in the training data (\cite{buhrmester2021analysis}). 

One infamous illustration of this bias is exemplified by the ``wolf in the snow" problem (\cite{lime}),
%GG: add citation (Original LIME paper had this problem mentioned)
where convolutional neural networks (CNNs) erroneously identify a husky as a wolf due to the presence of snow in the background. This happened because they learnt to associate `snow' with `wolf' based on the training data. This vulnerability to undesirable bias becomes particularly significant when these models are used in high-stakes settings such as disease diagnosis (\cite{CNN_medical1}) and autonomous vehicle operation (\cite{CNN_autonomous2}). In such scenarios, the consequences could be dire if, for example, a model incorrectly identifies a malignant condition as benign, or erroneously suggests it is safe to switch lanes into oncoming traffic due to misleading patterns in the training data.
% that become the model's biases, and chooses action to be taken in a critical scenario accordingly the consequences can be severe.
%where there is a chance of the model classifying a malignant case as benign and the action to be taken in a critical scenario respectively, based on spurious patterns in the data that become the models biases. \nat{Had to rephrase}
Such instances underscore the urgent need to address the bias-correction in CNNs to ensure their reliability and fairness in real-world applications. 

Recent works have shown that it is possible to obtain the knowledge of a trained CNN in the form of a symbolic rule-set, more specifically as a stratified Answer Set Program  (\cite{nesyfold, nesyfold-g}). The authors proposed a framework called NeSyFOLD, wherein the activation of filters in the final convolutional layer of the CNN serves as the truth value of predicates in the generated rule-set, offering valuable insights into the concepts learnt by the model and their relation to the target class to be predicted. The filters in the CNN are $n \times n$ real-valued matrices. These matrices capture the representation of various concepts in the images. The predicates are labelled as the concept(s) that their corresponding filters learn to identify in the images. Fig. \ref{fig_1} illustrates the NeSyFOLD framework and the final rule-set that is generated for a train set containing images of the `bathroom', `bedroom' and `kitchen' classes.

It is easy to scrutinize the rule-set generated by NeSyFOLD and find the biases that the CNN develops towards each class. The CNN's filters, during training, learn the most appropriate concepts that would easily distinguish between the different classes of images in the train set. To a human, a distinction made based on concepts found by the CNN may be counter-intuitive. Moreover, the concepts learned may only be adequate for differentiating between the classes of images present in the current train set. Frequently, these concepts might not suffice for accurately classifying images of the same class sourced differently, where key patterns in the images may vary. For instance, images from the new source might be captured from a different angle, at a different time of day, or under varying weather conditions. Hence, a human with adequate domain knowledge can identify the biases that are `undesired' or `desired' such that after retraining, the model becomes more robust to differently sourced data. An example of this could be a doctor identifying the concepts that appear in the generated rule-set, that are positively linked to the target class `malignant' and then suggesting the undesired and desired concepts. Those concepts if unlearnt/learnt by the CNN filters can improve the performance of the CNN on classifying images from different sources when deployed.

We introduce the NeSyBiCor (\underline{Ne}uro-\underline{Sy}mbolic \underline{Bi}as \underline{Co}rrection) framework, to aid in correcting pre-identified biases of the CNN. The ASP rule-set generated by NeSyFOLD, from the bias-corrected CNN serves to validate the effectiveness of the framework. The pre-identified biases are presented as semantic constraints based on concepts that should and should not be used to predict the class of an image. These concepts can be selected by scrutinizing the rule-set generated by NeSyFOLD.
% Building upon the foundation laid by NeSyFOLD, our approach leverages the rule-set generated by this framework to identify undesirable/desirable concepts and unlearn the undesirable concepts while reinforcing the desirable ones.
% The CNN is then retrained wAt the core of the NeSyBiCor framework is the novel semantic similarity loss function. 
We map the undesirable/desirable semantic concepts to their corresponding vector representations learnt by the filters. Next, we retrain the CNN with a novel semantic similarity loss function which is a core component of our framework. The loss function is designed to minimize the similarity of the representations learnt by the filters with the undesirable concepts and maximize the similarity to the desirable ones. Once the retraining is complete, we use the NeSyFOLD framework again to extract the refined rule-set. Hence, our approach provides a way to refine a given ASP rule-set subject to some semantic constraints.

To summarize, our contributions are as follows:
\begin{enumerate}
    \item We propose a novel framework, NeSyBiCor, for targeted bias correction in a CNN.
    \item We introduce a semantic similarity loss for penalizing/reinforcing filters that learn undesirable/desirable concept representations.
    \item We evaluate the framework on subsets of the Places (\cite{places}) dataset.
\end{enumerate}

\section{Background}
\subsection{Convolutional Neural Networks}
Convolutional Neural Networks (CNNs) are a sub-category of neural networks (NNs) well suited to pattern recognition in visual data first introduced by~\cite{lecun1989backpropagation}. Their distinctive feature is the presence of \emph{convolutional layers} which employ learnable filters (also called kernels) to extract spatial hierarchies of features from input data. 
% In its generic form, a convolution is a specialised linear operation on two functions with the same real-valued argument;
These filters are designed to detect specific features, such as edges or textures, by applying a mathematical operation called convolution. This operation involves sliding the filter over the image and computing the dot product of the filter values and the original pixel values in the image. The result of this process is a feature map, which is a new representation of the image emphasizing the detected features. Each filter in a layer can produce a distinct feature map, collectively forming a complex representation of the input that assists the network in learning to classify images or recognize patterns efficiently.
% The output of each filter as it performs the convolution operation over the input image is a feature map which is a 
% In CNNs, the inputs are the images, that are transformed by the learning algorithm and the filter while the output is a feature map (\cite{Goodfellow-et-al-2016}). A single feature map represents the response of a specific filter to the tensor - given a different filter the feature map corresponds to a different pattern in input data. An analysis of which types of inputs stimulate particular feature maps can be used to observe the decision-process of the CNN and its evolution in training (\cite{zeiler2014visualizing}).
%In its most generic form convolution is an operation on two functions with real-valued arument - in ML the first input is typically a multi-dimentional array of data and the second is the kernel. 
%\parth{Maybe need to add some info about the convolution operation of the filters to produce feature maps from input images and how their output can be treated as the representation of the pattern that they have learnt}
%\nat{Is that sort of what you wanted here, or something more mathematical? Not sure if you intend to refer back to something mathematical later or not.}
It has been shown that when a CNN is trained with images, the filters in its convolutional layers learn specific patterns. Moreover, the last layer filters learn to recognize high-level features such as objects or object parts. This has been used to increase the explainability of CNNs via analysis of the types of emergent patterns and the relations between them (\cite{zhang2018interpreting,zhang2017growing}). 
% as well as training CNNs to associate patterns with word phrases (\cite{wickramanayake2021comprehensible}).
In this work, we build upon the line of research where CNN filters are used as symbolic atoms/predicates in a rule-set for image classification (\cite{eric, nesyfold, nesyfold-g}).
\subsection{FOLD-SE-M and NeSyFOLD}
\medskip\noindent\textbf{FOLD-SE-M:}
Default logic is a non-monotonic logic used to formalize commonsense reasoning. A default $D$ is expressed as:
  
\begin{equation}\label{eq_1}
    D = \frac{A: \textbf{M} B}{\Gamma}
\end{equation}

\noindent Equation \ref{eq_1} states that the conclusion $\Gamma$ can be inferred if pre-requisite $A$ holds and $B$ is justified. $\textbf{M} B$ stands for ``it is consistent to believe $B$".
Normal logic programs can encode a default theory quite elegantly (\cite{gelfondkahl}). A default of the form: 
$$\frac{\alpha_1 \land \alpha_2\land\dots\land\alpha_n: \textbf{M} \lnot \beta_1, \textbf{M} \lnot\beta_2\dots\textbf{M}\lnot\beta_m}{\gamma}$$
\noindent can be formalized as the
normal logic programming rule:
$$\gamma ~\texttt{:-}~ \alpha_1, \alpha_2, \dots, \alpha_n, \texttt{not}~ \beta_1, \texttt{not}~ \beta_2, \dots, \texttt{not}~ \beta_m.$$
\noindent where $\alpha$'s and $\beta$'s are positive predicates and \texttt{not} represents negation-as-failure. We call such rules \emph{default rules}. 
Thus, the default 

$$\frac{bird(X): M \lnot penguin(X)}{flies(X)}$$

\noindent will be represented as the following default rule in normal logic programming:

{\tt flies(X) :- bird(X), not penguin(X).}

\noindent We call {\tt bird(X)}, the condition that allows us to jump to the default conclusion that {\tt X} flies, the {\it default part} of the rule, and {\tt not penguin(X)} the \textit{exception part} of the rule. 

FOLD-SE-M (\cite{foldsem}) is a Rule Based Machine Learning (RBML) algorithm. It generates a rule-set from tabular data, comprising rules in the form described above. The complete rule-set can be viewed as a stratified answer set program (a stratified ASP rule-set has no cycles through negation (\cite{Baral})). It uses special {\tt abx} predicates to represent the exception part of a rule where {\tt x} is a unique numerical identifier.   
FOLD-SE-M incrementally generates literals for \textit{default rules} that cover positive examples while avoiding covering negative examples. It then swaps the positive and negative examples and calls itself recursively to learn exceptions to the default when there are still negative examples falsely covered.

There are $2$ tunable hyperparameters, $ratio$, and $tail$. 
The $ratio$ controls the upper bound on the number of false positives to the number of true positives implied by the default part of a rule. The $tail$ controls the limit of the minimum number of training examples a rule can cover.
% FOLD-SE-M generates a much smaller number of rules than a decision-tree classifier and gives higher accuracy in general.

\medskip\noindent\textbf{NeSyFOLD:}
\begin{figure}[t]
    \centering
    \includegraphics[width=\linewidth, height = 8cm]{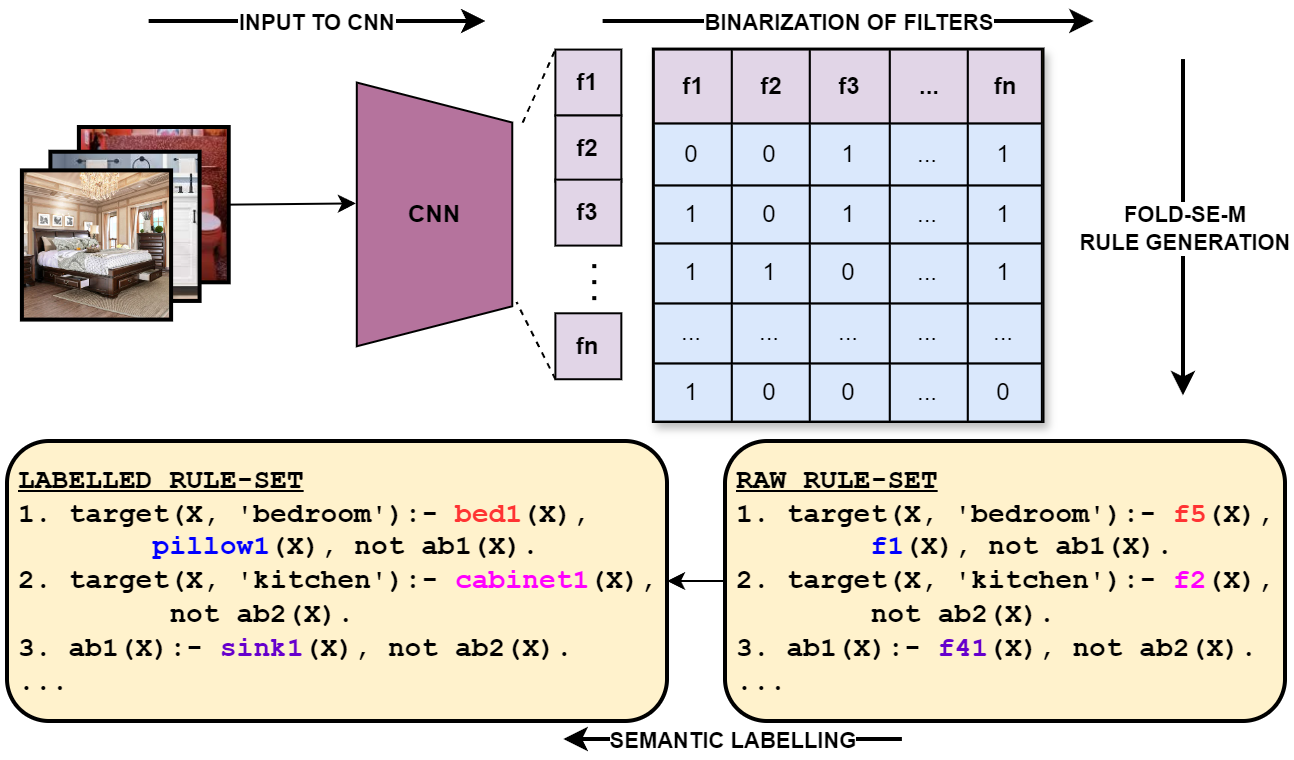}
    \caption{The NeSyFOLD Framework}
    \label{fig_1}
\end{figure}
\cite{nesyfold} introduced a neurosymbolic framework called NeSyFOLD to extract a symbolic rule-set in the form of a stratified Answer Set Program from the last layer filters of a CNN. Fig. \ref{fig_1} illustrates the NeSyFOLD framework.

\noindent\underline{\textit{Binarization of filter outputs:}} For each image, the output feature maps of the last layer filters in the CNN are collected and stored. Next, the norms of the feature maps are computed. A norm of a matrix here is simply adding the squared values of each element in the matrix and taking a square root of the output. Eq. \ref{eq_2} depicts this operation for the feature map produced by the $k^{th}$ filter for the $i^{th}$ image. This creates a table of norms of dimensions equal to \textit{no. of images} $\times$ \textit{no. of filters}. Finally, these norm values for each filter are treated as the filter's activation value and a weighted sum of the mean and standard deviation (Eq. \ref{eq_3}) of these values over all the rows (images) in the table determines the threshold of activation for each filter. $\alpha$ and $\gamma$ are hyperparameters in Eq. \ref{eq_3} and $n$ is the number of images in the train-set. $\theta_k$ is the threshold for the $k^{th}$ filter. This threshold is used to binarize each value. This creates a binarized vector representation of the image with a dimension equal to the number of filters in the last layer. Such vectors are computed for all images in the training set and collected in a binarization table (Fig.\ref{fig_1} top-right). Next, the FOLD-SE-M algorithm takes the binarization table as input and outputs a raw rule-set (Fig. \ref{fig_1} bottom-right) wherein each predicate represents a filter in the CNN's last layer. The truth value of each predicate is determined by the binarized activation of the corresponding filter in the CNN when classifying a new image using the rule-set.
\begin{align}
    a_{i,k} =& ~||A_{i,k}||_2 \label{eq_2}\\
    \theta_{k} =& ~\alpha \cdot \overline{a_{k}} + \gamma\sqrt{\frac{1}{n}\sum(a_{i,k} - \overline{a_{k}})^2} \label{eq_3}
\end{align}

Note that in the scope of our work, the abx/1 predicates (e.g. {\tt ab1(X)}, {\tt ab2(X)}, etc.) do not carry inherent semantic meaning but are crucial for the structural integrity and compact representation of the rule-sets generated. Each abx/1 predicate is found in the head of precisely one rule, and the corresponding rule body may contain semantically meaningful predicates linked to CNN filters or other auxiliary predicates. In essence, each abx/1 predicate in the rule-sets can be represented through a combination of conjunctions and disjunctions of predicates that possess semantic significance and are connected to the CNN’s filters. They are essential in simplifying the structure of the rule-sets.

\begin{figure}[h!]
    \centering
    \includegraphics[width=\linewidth]{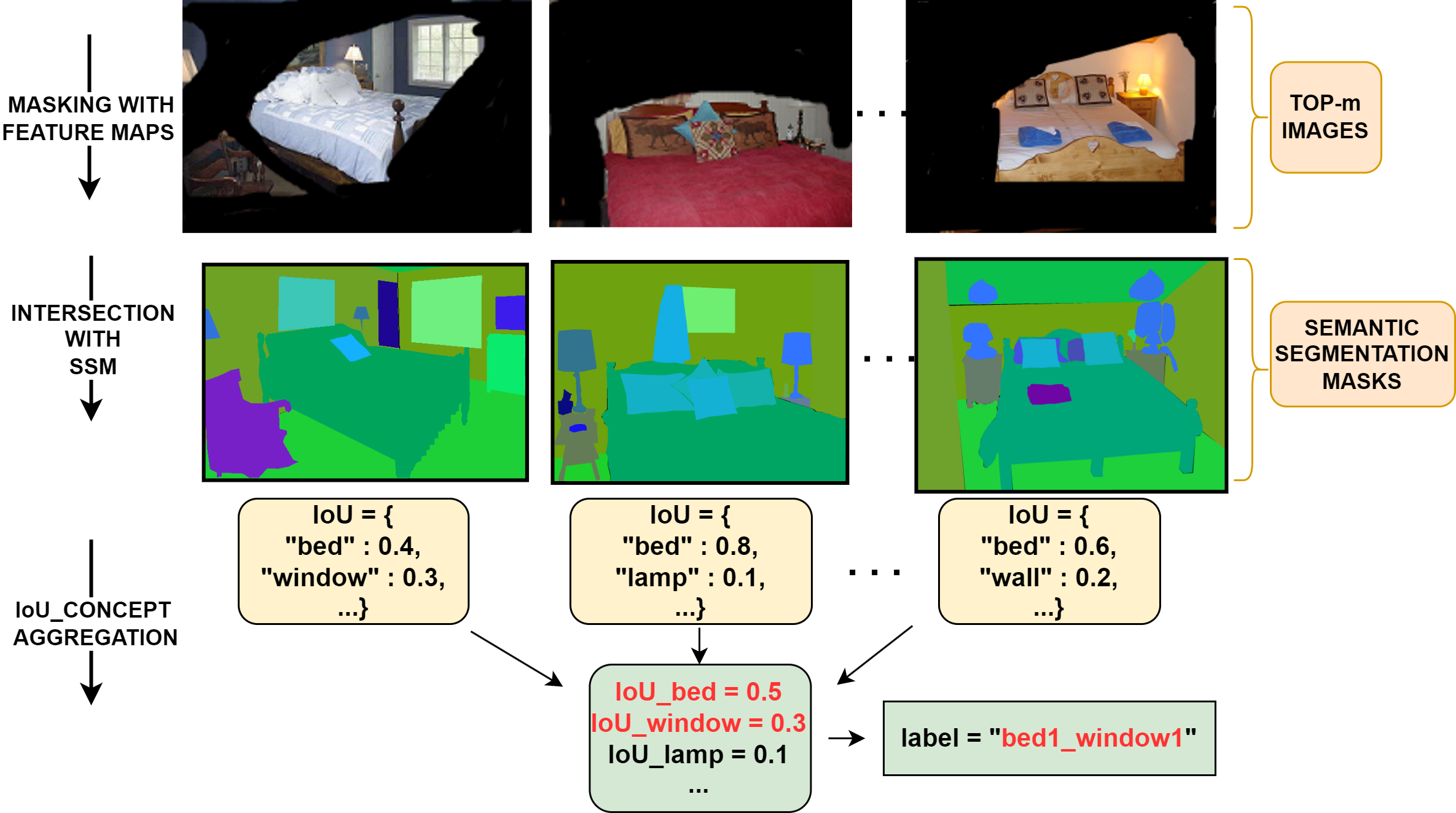}
    \caption{Semantic labelling of a predicate}
    \label{fig_sl}
\end{figure}

\noindent\underline{\textit{Semantic labelling of predicates:}}
Semantic segmentation masks of the images are used to map each filter to a set of concepts that it represents and thus the predicates in the rule-set can be given a semantic meaning. These semantic segmentation masks can be human annotated or those generated by using large foundation models such as SegGPT (\cite{seggpt}) or RAM (\cite{RAM}) in conjunction with the Segment Anything Model (SAM) (\cite{sam}). Hence, the rule-set serves as a highly interpretable global explanation of the CNN's decision making process. The predicted class of the image is represented in terms of logical rules based on concepts that the CNN filters learn individually. Fig. \ref{fig_sl} illustrates the semantic labelling of a single predicate in the raw rule-set extracted by the FOLD-SE-M algorithm or, in other words, it maps the predicate in the raw rule-set with semantic concept(s) that its corresponding filter has learnt. 
The filters in the last convolutional layer are known to learn high-level concepts, such as objects or object parts \cite{object_detectors_emerge}. First, the feature maps generated by a filter in the last convolutional layer for the \textit{top-m} images that activate it are collected. The \textit{top-m} images are selected according to the norm values of the feature maps generated by the filter for all images in the train set. Next, the feature maps are resized and masked onto the \textit{top-m} images (Fig. \ref{fig_sl} top). Notice that by doing this, one can observe the concepts that the filter is looking at in each of selected images. For example, the filter considered in Fig. \ref{fig_sl} is looking at beds in images.
The masked images are then overlapped onto the semantic segmentation masks and an Intersection over Union (IoU) score is calculated for each concept that is visible after the overlap (Fig. \ref{fig_sl}, middle).
Finally, the scores are aggregated over the \textit{top-m} images and the label of the predicate that is associated with the filter under consideration is determined (Fig. \ref{fig_sl}, bottom).
We refer the reader to the NeSyFOLD paper by \cite{nesyfold} for more details on the semantic labelling procedure and the NeSyFOLD framework itself.

% One of the applications of the NeSyFOLD framework is interpretable breast cancer classification. A CNN can be trained with breast cancer images and a interpretable rule-set can be extracted from the last layer filters. In case of semantic segmentation masks of the images being unavailable, the filter feature maps can be visualized to understand what regions of the image is each filter looking at.
% With the advancement in foundation models for vision, it has become possible to obtain semantic segmentation masks for images using tools such as SegGPT and SAM, improving the feasibility of NeSyFOLD. The inherent biases learnt by the CNN from the train data can also be identified by scrutinizing the rule-set. The filters learning the concepts/patterns in the images that are mere coincidental correlations w.r.t. the target class can be penalized using a cosine similarity based semantic similarity loss, hence correcting unwanted biases of the CNN and obtaining a more intuitive and generalized rule-set.
\section{Methodology}
% Should we name the framework? I think a name for the framework would be useful.
\begin{figure}[t]
    \centering
    \includegraphics[width=\linewidth, height = 9cm]{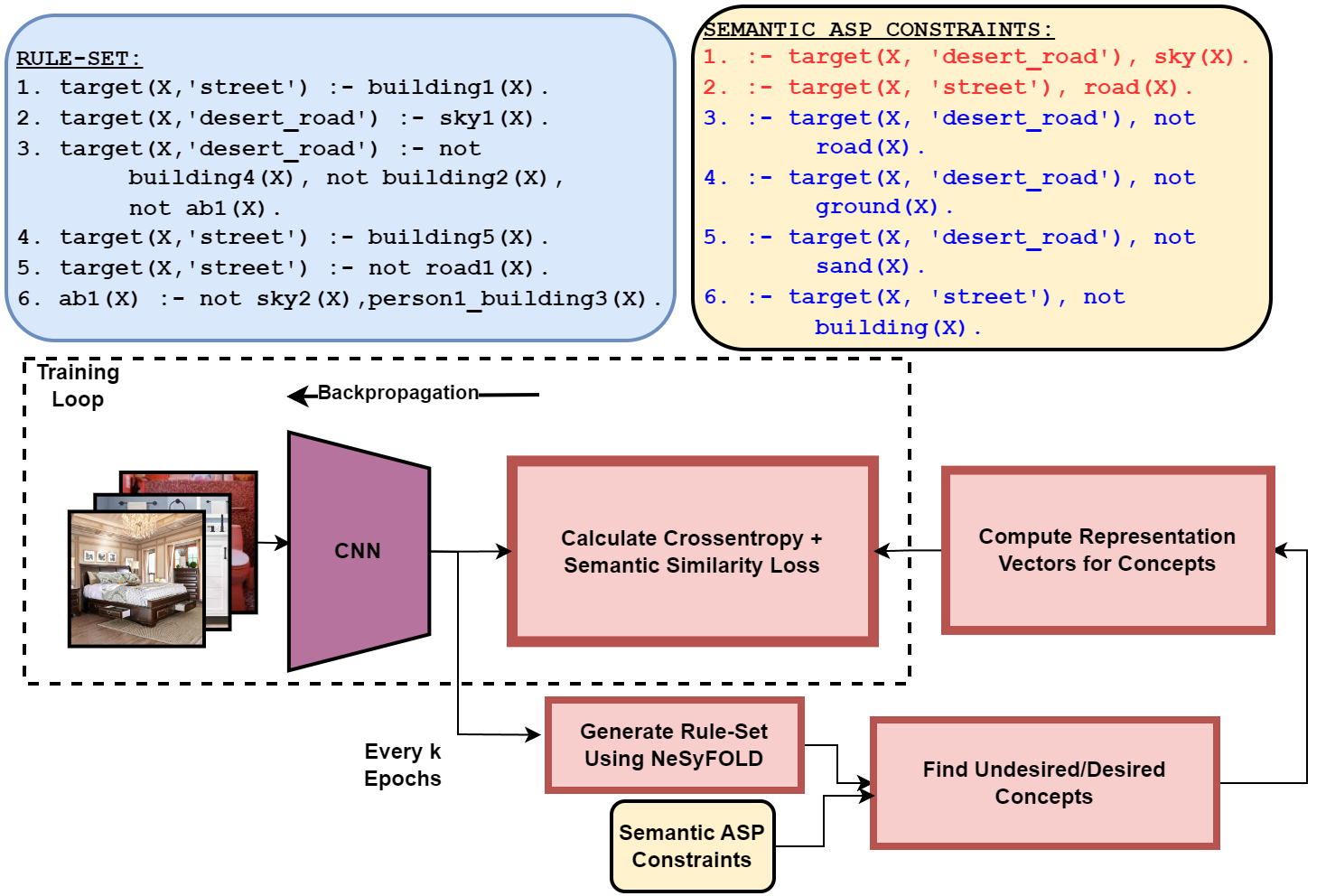}
    \caption{The NeSyBiCor Framework. Note that the crossentropy loss is calculated after the fully connected layer while the semantic similarity loss is calculated by using the filter output feature maps of the last convolution layer}
    \label{fig_2}
\end{figure}
We demonstrate the utility of our framework using classes of images from the Places dataset (\cite{places}). The dataset contains images of various indoor and outdoor places and scenes such as ``kitchen", ``bathroom", ``beach", ``forest", etc. We choose this dataset because the manually annotated semantic segmentation masks of the images are also readily available as part of a different dataset called ADE20k (\cite{ade20k}). The NeSyFOLD framework requires semantic segmentation masks for generating a meaningful rule-set with concept(s) as predicate names. 

In Fig. \ref{fig_2} we show a schematic diagram of the NeSyBiCor framework. As a running example, let us consider a CNN trained on $2$ classes of the places dataset namely `desert road' and `street'. NeSyFOLD generates a rule set as shown in the blue box at the top left of Fig. \ref{fig_2}. First, the desired and undesired biases are conceptualized in terms of the concepts that are desired or undesired to appear as being positively associated with a target class namely `desert road' or `street'. Notice that in rule 2, the `desert road' class is being predicted based on whether there is \textit{sky} visible in the image. Let us say this is an undesired concept for classifying images of `desert road' as is the \textit{road} concept for the `street' class. Note that `street' images are not being classified by the \textit{road} concept, meaning that the \textit{road} concept is not positively linked to the `street' class. Also note, that rule 5 has a predicate based on the \textit{road} concept (road1/1) but it appears with a negation preceding it hence it is not positively linked to the `street' class. However, it could happen that during our retraining the CNN might pick up on this concept so we can provide the undesired concepts beforehand. Similarly, the desired concepts are also identified by scrutinizing the rule-set or through domain knowledge.
The yellow box on the top-right of Fig. \ref{fig_2} shows the Semantic ASP constraints that are constructed for each class with the undesired (red) and desired (blue) concepts. The objective is to convert the given ASP rule-set to a rule-set that satisfies the given constraints as closely as possible.

Since the ASP rule-set is generated from a CNN, the intrinsic connectionist knowledge of the CNN needs to be revised based on the symbolic semantic constraints. Hence, we developed the following procedure to convert the semantic concepts that are meaningful to the human, to vector representations, meaningful to the CNN to facilitate retraining.

\medskip\noindent\textbf{Computing Concept Representation Vectors:}
The first step is to obtain the \textit{concept representation vectors} for each desired and undesired concept specified in the semantic constraints for each target class. Recall that a filter in the CNN is a matrix that detects patterns in the input image. For example, the filter associated with the predicate `sky1/1' in the rule-set is detecting some type of patterns in the sky (say, blue clear sky) and the filter associated with the `sky2/1' predicate is detecting some other types of patterns in the sky (say, evening sky) in the input `desert road' images. The output produced by a filter for a given image is again a matrix that can be flattened into a vector and treated as the representation of the patterns it has learnt to detect. Hence, to find the concept representation vector of  \textit{sky}, we first compute the individual \textit{filter representation vectors} for all the predicates that have the concept \textit{sky} in their name and are positively associated with the class `desert road'. In this case, there are only two, i.e., `sky1/1' and `sky2/1'. To compute their respective filter representation vectors, we find the top-10 images (inspired by NeSyFOLD) in the train-set that these filters are most activated by and take the mean of all $10$ vectors produced as their outputs for these $10$ images. Finally, to compute the concept representation vector for the concept \textit{sky}, we simply take the mean of the `sky1/1' and `sky2/1' filter representation vectors.

Repeating the above procedure for every undesired and desired concept yields their respective concept representation vectors.

Note that, as the \textit{number of images} in the train set increases, the time to find the top-10 images increases linearly. Hence, if the number of images in the train-set is doubled then the time taken to find the top-10 images would also nearly double.
In contrast, as the \textit{number of concepts} increases, the time taken to compute all the concept representation vectors depends largely on the number of filters that are associated with different concepts. If there are desired/undesired concepts that have no filters associated with them in the rule-set, then no concept representation vectors are computed. Thus, adding more desirable/undesirable concepts does not always translate to greater computation time.

Next, we start the retraining of the CNN with the original cross-entropy loss $\mathcal{L_{CE}}$ and our novel Semantic Similarity loss $\mathcal{L_{SS}}$. 
% as the representation vector for the respective filter concept produced as vector of all thethe aggregated representation vector can be calculated 

% has information about its learnt patterns, can be treated as the representation  feature maps produced by the filters can be used as a rthe various desired and undesired concepts that are positively associated with a target class in the rule-set.
% For example, ``sky1" and ``sky2" are two predicates whose corresponding filters in the CNN are capturing different patterns of the concept ``sky". Both of these predicates are positively associated with the class ``desert\_road". Hence, to find the rep

\medskip\noindent\textbf{Calculating the Semantic Similarity Loss:}
We define the semantic similarity loss $\mathcal{L_{SS}}$, for a train set with $N$ images and CNN with $K$ filters in the last convolutional layer, as follows:

\begin{align}
    \mathcal{L_{SS}} = \sum_{i = 1}^N\left[\sum_{j = 1}^{K}\left[\lambda_B\sum_{b \in \mathbf{B}}cos\_sim(\mathbf{r}_j^i, \mathbf{r}_b) - \lambda_G\sum_{g \in \mathbf{G}}cos\_sim(\mathbf{r}_j^i, \mathbf{r}_g)\right]\right]
\end{align}

The $cos\_sim$ function calculates the cosine similarity between two representation vectors. $\mathbf{r}_j^i$ is the filter representation vector obtained from the $i^{th}$ image's $j^{th}$ filter output. $\mathbf{r}_b$ is the concept representation vector for some concept $b$ in the list of undesired concepts $\mathbf{B}$. Similarly, $\mathbf{r}_g$ is the concept representation vector for some concept $g$ in the list of desired concepts $\mathbf{G}$. $\lambda_B$ and $\lambda_G$ are hyperparameters. 

The rationale behind the loss function is simple: the loss increases when the filter representation vectors closely resemble the undesired concept representation vectors and decreases when the filter representation vectors are closer to the desirable concept representation vectors. This approach is conceptually similar to the loss function used in word2vec (\cite{word2vec}), where the model's objective is to maximize the similarity between a target word and its context words while minimizing similarity with randomly sampled words.
Thus, as training progresses with any standard optimization technique such as Adam (\cite{adamoptim}), the loss is minimized and the filters get pushed away from learning undesirable concepts and pushed towards learning desirable concepts. Note that the total loss is defined as $\mathcal{L_{TOTAL}} = \mathcal{L_{CE}} + \mathcal{L_{SS}}$. Hence, the cross-entropy loss $\mathcal{L_{CE}}$ is also jointly minimized along with the semantic similarity loss $\mathcal{L_{SS}}$,  to maintain the classification performance.

An important observation to make here is that the semantic similarity loss might become negative as the first term tends to $0$. The total loss is the sum of the semantic similarity loss and the cross-entropy loss. The hyperparameters $\lambda_B$ and $\lambda_G$ help to determine the influence of the semantic similarity loss towards the total loss. This is common practice in machine learning literature to control the influence of various loss terms in the total loss. Hence even if the semantic similarity loss is negative, due to the hyperparameters which are tuned on a validation set, the cross-entropy loss is not highly influenced which helps in maintaining the model’s classification performance. This is demonstrated by our experiments later in the paper.

\medskip\noindent\textbf{Recalibrating the Concept Representation Vectors: }
As the training progresses, the CNN filters may learn slightly different representations of undesirable/desirable concepts. For example, some filters might pick up a different pattern in the sky of the images that was not caught earlier when the training started. Now if this new pattern in the sky (say, dense clouds) is again deemed sufficient by the CNN to be a significant feature in distinguishing the `desert road' class from the `street' class, then this filter would show up in the rule-set as a `skyx/1' predicate.
This would happen because the initially computed concept representation vector for \textit{sky} does not encapsulate the representation for this new type of \textit{sky} which became significant because the filters were pushed away from learning the other types of \textit{sky} by the loss function.
It is also possible that some new undesirable concept such as \textit{road} for the `street' class shows up in the rule-set. In such a case, the new concept would not be mitigated as there was no filter capturing the representations for \textit{road} initially. Hence, no concept representation vector for \textit{road} is available to push the filters away from learning this new undesirable concept. Similarly, some new filters, learning known or new desirable concepts might also appear. 
% If these new representations could be used to hasten the training process and pull the filters faster towards a broader variety of desirable concept representations. 

To solve this problem, we propose rectifying all the concept representation vectors for each class after every $k$ epochs during training. We do this by running the NeSyFOLD framework after every $k$ epochs and obtaining a new rule set from the partially retrained CNN. We then compute the concept representation vectors for all the undesired and desired concepts again by considering all the predicates that appear in the newly generated rule-set, by following the algorithm described above.
We then aggregate the new concept representation vectors with the old concept representation vectors by taking their mean, so that the new aggregated vectors encapsulate the information of the new patterns that were found. Hence, the filters can now be pushed away from/towards these new undesirable or desirable representations respectively. This way we can ensure that at the end of retraining, there is a greater chance that the new ASP rule-set produced from the CNN satisfies the semantic constraints posed against the initial rule-set.

% So given a concept (e.g. \textit{sky}) all the predicates in the rule-set that have `sky' in their label (e.g. sky1/1, sky2/1 etc.) are selected. Each predicate's truth value is linked to a particular filter in the CNN's last convolution layer. Hence the corresponding filters (e.g. sky1/1 corresponds to filter no. $234$, sky2/1 corresponds to filter no. $31$ etc.) are found and the top-10 images that activate them the most are found from the train set. Therefore, as the \textit{number of images} in the train set increases, the time to find the top-10 images increases linearly. Hence, if the number of images in the train-set is doubled then the time taken to find the top-10 images would also nearly double.

As the \textit{number of concepts} increases, the time taken to compute all the concept representation vectors depends largely on the number of filters that are associated with different concepts. If there are desired/undesired concepts that have no filters associated with them in the rule-set, then no concept representation vectors are computed. Thus, adding more desirable/undesirable concepts does not always translate to greater computation time.
\section{Experiments}
We conducted experiments to address the following research questions:

\medskip\noindent\textbf{Q1:} How does our NeSyBiCor framework affect the initial rule-set?

\medskip\noindent\textbf{Q2:} How does the performance of the rule-set extracted from the bias-corrected CNN compare against the one extracted from vanilla CNN w.r.t. accuracy, fidelity and size?

\medskip\noindent\textbf{Q3:} What percentage of the covered examples are classified using undesired/desired concepts before and after applying the NeSyBiCor framework?

\medskip\noindent\textbf{[Q1] Bias Corrected Rule-Set:}
The core idea of correcting the bias of the CNN and by extension, the rule-set, is that the images should be classified by rules that use concepts that are intuitive to humans and might be more representative of the class to which the image belongs. Hence, given constraints on the undesired and desired concepts to be learnt by the CNN, the ideal outcome should be that the undesired concepts are eradicated and desired concepts are infused into the revised rule-set.

\noindent\textbf{Setup}
We train a CNN on subsets of the Places (\cite{places}) dataset.
We selected 3 subsets of 3 classes i.e. \textit{babek} (`\underline{ba}throom', `\underline{be}droom', `\underline{k}itchen'), \textit{defs} (`\underline{de}sert road', `\underline{f}orest road', `\underline{s}treet') and \textit{dedrh} (`\underline{de}sert road', `\underline{dr}iveway', `\underline{h}ighway') along with 3 subsets of 2 classes i.e. \textit{babe} (`\underline{ba}throom', `\underline{be}droom'), \textit{des} (`\underline{de}sert road', `\underline{s}treet') and \textit{deh} (`\underline{de}sert road', `\underline{h}ighway').
We employed a VGG16 CNN, pre-trained on \textit{Imagenet} (\cite{deng2009imagenet}), training over $100$ epochs with batch size $32$. The Adam (\cite{adamoptim}) optimizer was used, accompanied by class weights to address data imbalance. $L2$ Regularization of $0.005$ spanning all layers, and a learning rate of $5 \times 10^{-7}$ was adopted. A decay factor of $0.5$ with a $10$-epoch patience was implemented. Images were resized to $224 \times 224$. Each class has $5000$ images so we used $4000$ as the train set and $1000$ as the test set per class.
Finally, we used NeSyFOLD to generate the rule-set.

Next, we manually identified the desired and undesired concepts for each class as shown in Table \ref{tb_1}. We then used the NeSyBiCor framework on all the subsets listed above with the appropriate semantic constraints based on each class's desired and undesired concepts. We used a value of $5e^{-2}$ and $1e^{-3}$ for the $\lambda_B$ and $\lambda_G$ hyperparameters (after empirical evaluation on a validation set) while computing the semantic similarity loss respectively. We retrained for $50$ epochs and recalibrated the concept representation vectors every $5$ epochs. NeSyFOLD was used on the retrained CNN to obtain the bias-corrected rule-sets for each subset.
\begin{table*}[t]
\fontsize{9}{10}\selectfont
\begin{tabular}{@{}rll@{}}
\toprule
Class & Undesired Concepts & Desired Concepts\\ \midrule
bathroom & wall, floor & sink, toilet\\ \cmidrule(lr){1-3}
bedroom & wall, floor & bed\\ \cmidrule(lr){1-3}
kitchen & wall, cabinet & work surface, range, sink\\ \cmidrule(lr){1-3}
desert road & sky, building & ground, sand, desert\\ \cmidrule(lr){1-3}
forest road & sky, building & tree\\ \cmidrule(lr){1-3}
street & sky, tree & building\\ \cmidrule(lr){1-3}
driveway & sky, building & house\\ \cmidrule(lr){1-3}
highway & sky, tree & road\\ \cmidrule(lr){1-3}

\end{tabular}
\caption{The desired and undesired concepts for the classes of images in Places dataset}
\label{tb_1}
\end{table*}

\begin{figure}[t]
    \centering
    \includegraphics[width=\linewidth, height = 9cm]{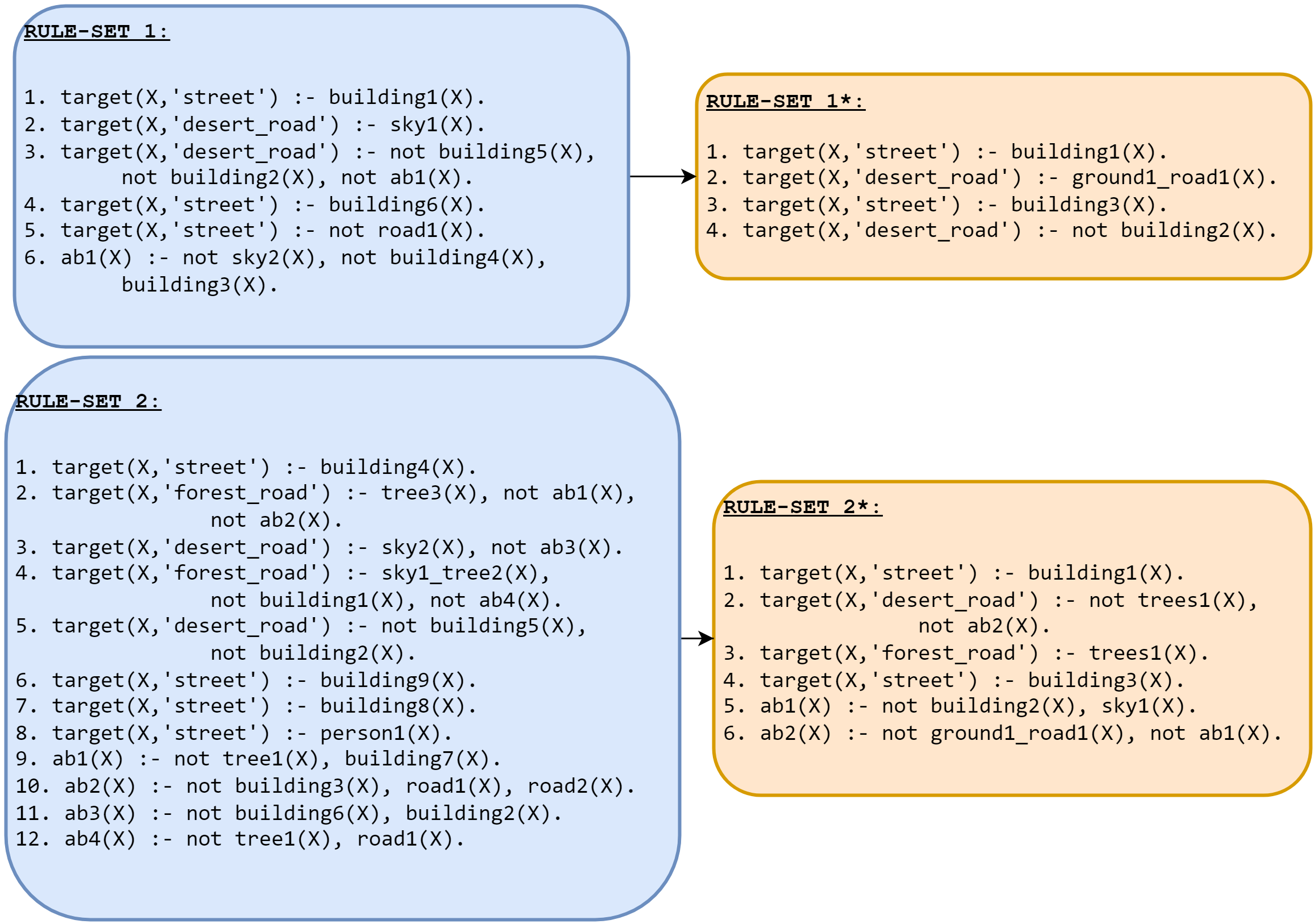}
    \caption{The initial and final rule-sets after applying the NeSyBiCor framework on the CNNs trained on \textit{des} (RULE-SET 1) and \textit{defs} (RULE-SET 2)}
    \label{fig_3}
\end{figure}

\noindent\textbf{Result:}
In Fig. \ref{fig_3} we show the initial (RULE-SET 1 \& RULE-SET 2) and final bias-corrected (RULE-SET 1* \& RULE-SET 2*) rule-sets for $2$ of the $6$ subsets, namely \textit{des} and \textit{defs}. Due to the lack of space, we present the other rule-sets in the supplementary material (\cite{nesybicor_arxiv}).

Recall that, the undesired concepts for the `desert road' class are `sky' and `building'. In RULE-SET 1, rule $2$ uses the `sky1/1' predicate to determine if the image belongs to the `desert road' class. In the bias-corrected rule-set, RULE-SET 1*, there is no `sky' based predicate. Moreover, the only predicate positively linked with the `desert road' class is the `ground1\_road1/1' predicate which is based on the desired concept `ground' and refers to the corresponding filter, learning a pattern comprising of specific type of patches of `ground' and `road'. Note, that the name of the predicate is determined by NeSyFOLD, which uses manually annotated semantic segmentation masks for the images. Ideally, it would be more appropriate for the predicate to be labelled as `sand1\_road1' but we are limited by the available annotations. One could generate images that are masked with the receptive field of the corresponding filter and manually label the predicate based on the concepts that are visible in the image as shown by \cite{eric}. 

Note that the rule-set is generated by the FOLD-SE-M algorithm from the binarized feature map outputs of the filters in the last convolutional layer of the CNN. Hence the rule-set mirrors the representations that are learnt by the CNN filters. As the re-training progresses, the filters are penalized for learning representations of the undesired concepts `sky' and `building', in case of the `desert road' class. Thus at the end of the bias correction, very few/none of the filters learn representations of the `sky' or `building' concepts. The filters tend to learn representations for the desired concepts (e.g. `ground') more. Thus naturally when the filters' feature maps are binarized and the binarization table is obtained, it is the filters that have learnt these desired concepts that form better features and are selected by the FOLD-SE-M algorithm to appear in the rule-set. Hence the undesired concepts appear to have been “dropped” by the algorithm in the bias-corrected rule-set.

Similarly, looking at RULE-SET 2* it is apparent that the number of undesired predicates associated positively with a class is reduced w.r.t. RULE-SET 2. Also, notice that certain predicates that are capturing irrelevant concepts such as `person' are eradicated.
It is clear by simply examining the rule-sets that in general, the rule-sets are more in line with the specified semantic constraints. Hence, the predictions being made are based on the concepts that the human finds more intuitive.

\begin{table}[t]
\fontsize{9}{10}\selectfont
\begin{tabular}{@{}rlllllll@{}}
\toprule
\multicolumn{1}{l}{Data} & Algo & Fid. & Acc.& Pred. & Size & \% Undesired & \% Desired \\ \midrule
\multirow{2}{*}{\textit{babek}}     & NeSyBiCor & $0.74$ & $0.78$ & $\textbf{7}$ & $\textbf{11}$ & $\textbf{0.47}$ & $\textbf{0.53}$\\
                        & Vanilla & $\textbf{0.87}$ & $\textbf{0.86}$ & $24$ & $41$ & $0.66$ & $0.33$\\ \cmidrule(lr){1-8}
\multirow{2}{*}{\textit{defs}}   & NeSyBiCor & $0.86$ & $\textbf{0.93}$ & $\textbf{6}$ & $\textbf{9}$ & $\textbf{0.01}$ & $\textbf{0.99}$\\
                         & Vanilla & $\textbf{0.93}$ & $0.91$ & $16$ & $23$ & $0.36$ & $0.57$\\\cmidrule(lr){1-8}

\multirow{2}{*}{\textit{dedrh}}      & NeSyBiCor & $0.78$ & $0.74$ & $\textbf{12}$ & $\textbf{24}$ & $\textbf{0.29}$ & $\textbf{0.57}$\\
                            & Vanilla & $\textbf{0.82}$ & $\textbf{0.79}$ & $30$ & $51$ & $0.67$ & $0.3$\\ \cmidrule(lr){1-8}
% \multirow{2}{*}{dehs}      & NF-BiCor & $0.74$ & $\textbf{0.84}$ & $\textbf{13}$ & $\textbf{22}$ & $0.19$ & $0.75$\\
%                             & NF-V & $\textbf{0.88}$ & $0.83$ & $29$ & $43$ & $\textbf{0.03}$ & $\textbf{0.85}$\\ \cmidrule(lr){1-8}
% \multirow{2}{*}{defh}      & NF-BiCor & $0.6$ & $\textbf{0.79}$ & $\textbf{9}$ & $\textbf{15}$ & $0.18$ & $0.48$\\
%                             & NF-V & $\textbf{0.83}$ & $0.76$ & $25$ & $41$ & $\textbf{0.05}$ & $\textbf{0.95}$\\
%                             \\ \cmidrule(lr){1-8}               
\multirow{2}{*}{\textit{babe}}      & NeSyBiCor & $0.88$ & $\textbf{0.93}$ & $\textbf{4}$ & $\textbf{6}$ & $\textbf{0.0}$ & $0.6$\\
                            & Vanilla & $\textbf{0.92}$ & $0.92$ & $14$ & $23$ & $0.11$ & $\textbf{0.82}$\\ \cmidrule(lr){1-8}
\multirow{2}{*}{\textit{des}}      & NeSyBiCor & $0.96$ & $\textbf{0.97}$ & $\textbf{4}$ & $\textbf{4}$ & $\textbf{0.0}$ & $\textbf{0.95}$\\
                            & Vanilla & $\textbf{0.97}$ & $0.97$ & $9$ & $10$ & $0.46$ & $0.47$\\\cmidrule(lr){1-8}
\multirow{2}{*}{\textit{deh}}      & NeSyBiCor & $0.77$ & $\textbf{0.86}$ & $\textbf{6}$ & $\textbf{6}$ & $\textbf{0.36}$ & $0.29$\\
                            & Vanilla & $\textbf{0.91}$ & $\textbf{0.86}$ & $18$ & $27$ & $0.46$ & $\textbf{0.47}$\\\cmidrule(lr){1-8}\\\cmidrule(lr){1-8}

\multirow{2}{*}{\textbf{MS}}      & NeSyBiCor & $0.83$ & $0.87$ & $\textbf{6.5}$ & $\textbf{10}$ & $\textbf{0.19}$ & $\textbf{0.66}$\\
                            & Vanilla & $\textbf{0.90}$ & $\textbf{0.89}$ & $18.5$ & $29.17$ & $0.45$ & $0.49$\\\cmidrule(lr){1-8}
                            
% \\\cmidrule(lr){1-6}                            
% \multirow{3}{*}{\textbf{MS}}      & ERIC & $0.70 \pm 0.01$ & $0.72 \pm 0.01$ & $71 \pm 4$ & $211 \pm 15$\\
%                             & NF   & $0.78 \pm 0.02$ &$0.75 \pm 0.02$ & $63 \pm 8$ & $120 \pm 19$ \\
%                             & NF-E & $ \textbf{0.80} \pm \textbf{0.03} $ &$ \textbf{0.78} \pm \textbf{0.03} $ & $ \textbf{20} \pm \textbf{4} $ & $ \textbf{36} \pm \textbf{9} $ \\ \cmidrule(lr){1-6}

\end{tabular}
\caption{Comparison between the rule-set generated before (Vanilla) and after (NeSyBiCor) bias correction with the NeSyBiCor framework. \textbf{MS} shows the average value for each evaluation metric. Bold values are better.}
\label{tb_2}
\end{table}

\medskip\noindent\textbf{[Q2, Q3] Qualitative evaluation:}
Recall that, the rule-set acts as the final decision making mechanism for any input image to the CNN. The input image is converted to a binarized vector of dimension equal to the number of filters in the last convolutional layer of the CNN. Each value in the binarized vector represents the activation (1) /deactivation (0) of the corresponding filter. Since each filter is mapped to a predicate in the rule-set, the truth value of the predicate is determined by the activation of the filter. 
The bias-corrected rule-set should be more faithful to the semantic constraints while sacrificing minimal accuracy, fidelity and interpretability.

\noindent\textbf{Setup:} We use the previously generated rule-sets for all subsets to classify the test images. In Table \ref{tb_2} we show the comparison between the accuracy (Acc.), fidelity (Fid.), number of unique predicates (Pred.) and rule-set size (Size) or total number of predicates between the initial and the bias-corrected rule-set for each of the $6$ subsets. The top three rows are $3$-class subsets and the bottom three are $2$-class subsets. We use rule-set size as a metric of interpretability. \cite{rulesetinterpretability} showed through human evaluations that as the size of the rule-set increases the difficulty in interpreting the rule-set also increases.
We also qualitatively evaluate how well the NeSyBiCor framework performs in eradicating the undesired concepts while introducing desired ones. Hence, for a given rule-set, out of all training images that were classified by the rule-set as any one of the classes, we report the fraction of images that followed a decision path that involved an undesired predicate. We report this value under the \% Undesired column in Table \ref{tb_2}. If there was no undesired predicate in the decision path then we check for a desired predicate. This value is reported as \% Desired in Table \ref{tb_2}. 

\noindent\textbf{Result:} The accuracy of the bias-corrected rule-set is comparable to the initial rule-set for all the $2$-class subsets. For the $3$-class subsets, the accuracy drops in 2 cases, \textit{babek} and \textit{dedrh} but is 
higher for \textit{defs} showing no clear trend. This might be because for a smaller number of classes, it is easier for the filters to learn alternate representations which is exactly what we are doing. This is expected, and as the number of classes increases, it will become difficult to optimize for accuracy as well as learning/unlearning concepts. A similar reason applies to lower fidelity values as well. Note that a loss of accuracy on the current data could mean a gain in accuracy in a dataset that is sourced differently hence making the rule-set more robust.

The number of unique predicates and rule-set size is consistently reduced in the bias-corrected rule-set with an average reduction of approximately $65\%$ in the number of unique predicates and rule-set size. Recall that within the NeSyFOLD framework, the FOLD-SE-M algorithm generates the final rule-set by processing binarized filter outputs of all training images, organized into a structure known as the binarization table (refer to Fig. \ref{fig_1}). This process is akin to how a decision tree algorithm selects the most significant features that effectively segment the majority of the data. A reduction in the number of unique predicates within the rule-set indicates that fewer features are required to successfully differentiate the data while preserving accuracy. This reduction suggests that during retraining, the filters are increasingly focusing on learning more targeted representations (ideally the desired ones), which are distinctly relevant for their respective target classes.
% An interesting observation is that the rule-set size and number of predicates in the bias-corrected rule-set is always lesser than the initial rule-set. This is because the semantic similarity loss encourages the CNN to focus on learning to distinguish between classes by learning only a few representations that are as different as possible from the representations of the undesired concepts. This indirectly encourages sparsity in CNN.

Finally, the most relevant observation is that the \% Undesired value is always lesser for the bias-corrected rule-set as expected with an average reduction of $58\%$. This means that there is a reduction in the number of images being classified by following a path in the decision-making that involved an undesired concept. Since we used a higher $\lambda_B$ than $\lambda_G$ while calculating the semantic similarity loss, the CNN focused more on learning representations that are dissimilar to the undesired concepts. The \% Desired value is higher in all cases except for \textit{babe} and \textit{deh} with an average increase of $35\%$. This means that the number of images being classified with at least one desirable concept in the decision path is increased. Note that in the case of \textit{babe} and \textit{deh}, a lower \% Desired value means that the images were being classified by some other concepts that the human is indifferent towards. In practical scenarios, the primary concern is to ensure images are not categorized based on undesired concepts, a capability effectively demonstrated by the NeSyBiCor framework in our experiments.

\section{Related Work}
The problem of bias in machine learning is well known and many efforts to mitigate it have been made in the last years as surveyed by \cite{mehrabi2021survey} who provide an in-depth overview of types of biases as well as bias mitigation techniques specific to different machine learning fields. 

When considering visual datasets and CNNs, there are multiple distinct types of bias as described by \cite{torralba2011unbiased}: selection bias, framing bias, label bias, and negative set bias. Selection bias occurs if the selection of subjects for the datasets differs systematically from the population of interest - for example, 
a dataset may lack the representation of a certain gender or ethnicity (\cite{buolamwini2018gender}). Framing bias refers to both the selection of angle and composition of the scene when taking a photo as well as any editing done on the images. 
% \cite{heuer2011obesity} discovered that in works on obesity images of people classed as obese were often captured without the head of the subject in the frame in contrast to non-obese subjects.
Label bias arises from errors in labelling the data in comparison to ground truth as well as poor semantic categories. The latter is more common; for example if in a dataset exist two classes - ``grass''
vs. ``lawn” - different labellers may assign different labels to the same image (\cite{malisiewicz2008recognition}). Lastly, the negative set bias refers to a \emph{negative} set which is what in a dataset is considered ``rest of the world'' (e.g. ``car'' and ``not-car''); if this set is unbalanced it can negatively affect the classifier (\cite{torralba2011unbiased}). 

There are several works on bias detection in CNNs and neural networks in general. \cite{cnn_representations_bias} propose one such method for detecting the learnt bias of a CNN. They show how spurious correlations can be detected by mining attribute relationships in the CNN without the use of any extra data. For example, the CNN might mistakenly learn the ``smiling" attribute to be correlated with the ``black hair" attribute due to bias in the dataset. Their method can help detect such correlations. \cite{ifbid} propose an approach to analyze the biases in neural networks by observing identifiable patterns in the weights of the model. Unlike our NeSyBiCor framework, these methods do not correct the identified biases.

%, as described by \cite{mehrabi2021survey} which can be broadly divided into three categories: Data to Algorithm, Algorithm to User and  User to Data. The first category encompasses biasaes present in the data that when used in ML, can result in biased models - such as representation bias, which arises from how the data is sampled in a population (\cite{suresh2021framework}). 
Earlier forms of bias mitigation involve transforming problems or data to address bias or imbalance, and over the years became more specialized resulting in algorithms that re-balance internal
distributions of training data to minimize bias (\cite{chakraborty2021bias}). Various new approaches have also been proposed that endeavor to tackle bias either during training or by modifying the already trained model. For example
\cite{zhao2017men} present a method based on Lagrangian relaxation for collective inference to reduce it. Many methods are generic and applicable across many types of ML models, such as the measure of decision boundary (un)fairness designed by \cite{zafar2017fairness} or convex fairness regularizers for regression problems as introduced by \cite{berk2017convex}. 

Logic-based bias mitigation is a comparatively smaller and newer area that offers promising results. In NLP \cite{cai2022mitigating} present a model based on neural logic, Soft Logic Enhanced Event
Temporal Reasoning (SLEER), which utilizes t-norm fuzzy logics to acquire unbiased temporal common sense (TCS) knowledge from text. Various methods of incorporating logical constraints to increase the fairness of the model, either before or after training, have also been proposed. Some of them are problem-specific, e.g. related to voting fairness (\cite{celismultiwinner}) others are more general (\cite{goh2016satisfying}). 
 
Similarities can also be seen in more general approaches to incorporating logical constraints in ML not explicitly related to bias mitigation - either during training e.g. using custom property-based loss functions (\cite{fischer2019dl2,slusarz2023logic}) or
%by various methods of 
modification of an existing model (\cite{leino2022self}).

\section{Conclusion and Future Work}
We proposed a neurosymbolic framework called NeSyBiCor for correcting the bias of a CNN. Our framework, in addition to correcting the bias of a CNN also allows the user to fine-tune the bias based on general concepts according to their specific needs or application. To the best of our knowledge, this is the first method that does bias correction by using the learnt representations of the CNN filters in a targeted manner. We show through our experiments that the bias-corrected rule-set is highly effective at avoiding the classification of images based on undesired concepts. It is also more likely to classify the images based on the desired concepts. The main component of the NeSyBiCor framework is the semantic similarity loss. It serves as a measure of similarity between the representations learnt by the filters, to the representation of the undesired and desired concepts. Another benefit of the NeSyBiCor framework is that the bias-corrected rule-set is smaller in size, thus improving the interpretability.

As part of future work we intend to use tools such as those designed by \cite{seggpt} and \cite{RAM}  that use vision foundation models for automatic semantic segmentation of images. Using these tools can help create semantic segmentation masks for images with custom names for the concepts that appear in the image. Currently, we are using pre-annotated semantic segmentation masks which often can be limited in the variety of concepts annotated, as well as their labels.
With regard to scalability, we plan to examine the effects of scaling to datasets with a large number of classes. The challenge here is that the rule-set generated by the NeSyFOLD framework that we employ tends to lose accuracy critically as the number of classes increases due to the binarization of the filter activations. 

Another factor that can diminish accuracy when utilizing the NeSyBiCor framework is the selection of inappropriate concepts or essential concepts that are crucial for the CNN to perform classifications effectively. For example, if the concept “bed” is deemed as undesired by the user, then the filters in the CNN will avoid learning the representations of beds in the images of bedrooms but since it is a significant feature of most bedroom images, there might not be other features that could help the CNN distinguish the bedroom images from other images hence hurting accuracy of the CNN.
We intend to investigate methods to maintain accuracy even if the selected concepts are essential.

Finally, the work we presented here may be used to extend implementations of loss functions based on Differentiable Logics:~\citeN{fischer2019dl2,slusarz2023logic,Dag24}. So far, the idea of compiling a loss function from an arbitrary logical formula (advocated in the above papers) was based on the assumption that during training, checking for property satisfaction is as easy as checking class labels. Therefore, logical loss functions can be easily combined with the standard cross-entropy loss function in backpropagation training.
For example, the famous property of \emph{neural network robustness} involves a trivial robustness decision procedure at the training time~(cf. \cite{CKDKKAE22}).
 This paper gives an example of a neuro-symbolic training scenario when a property of interest cannot be decided at the training time, and we had to resort to using CNN filters in order to replace a deterministic decision procedure. As the Differentiable Logic community continues to extend its range of applications (see e.g.~\cite{flinkow2024comparingdifferentiablelogicslearning}), solutions such as the ones presented here will be needed in order to approximate the decision procedures. We intend to investigate this line of research.

\bibliographystyle{tlplike}
\bibliography{bibliography}
\newpage
\appendix
% \centerline{\huge Appendix I}
\section{}
\noindent In the main paper we could only show the initial and bias-corrected rule-sets for two of the six chosen subsets of the Places dataset. Hence, here we present the pre bias correction rule-sets and the bias-corrected rule-sets for the other four chosen subsets. 

\subsection{\textit{babek} (`bathroom', `bedroom', `kitchen'):}
\medskip\noindent\textbf{Initial:}\\
\small{
\begin{verbatim}
target(X,`kitchen') :- cabinet1_wall1(X), not ab4(X), not ab5(X).
target(X,`bedroom') :- bed5(X), not ab6(X).
target(X,`bathroom') :- sink1_toilet1_wall4(X).
target(X,`bathroom') :- not cabinet4(X), wall10(X), not cabinet9_wall9(X).
target(X,`bedroom') :- bed9(X), not ab7(X), not ab9(X).
target(X,`kitchen') :- cabinet3(X).
target(X,`bathroom') :- sink2_wall5(X).
target(X,`bathroom') :- not bed1(X), not bed6(X), wall7_sink3(X).
target(X,`bedroom') :- bed4(X).
target(X,`kitchen') :- cabinet9_wall9(X).
target(X,`bathroom') :- not wall6(X), not bed7(X).
ab1(X) :- not bed6(X), cabinet3(X).
ab2(X) :- bed3(X), not ab1(X).
ab3(X) :- cabinet9_wall9(X), not ab2(X).
ab4(X) :- not cabinet5(X), not ab3(X).
ab5(X) :- not cabinet7(X), not cabinet6_wall8(X), sink1_toilet1_wall4(X).
ab6(X) :- wall7_sink3(X), sink1_toilet1_wall4(X).
ab7(X) :- not bed8(X), wall2(X).
ab8(X) :- not cabinet8(X), not cabinet2_wall3(X).
ab9(X) :- not bed2(X), not bed3(X), not ab8(X).
\end{verbatim}
}

\medskip\noindent\textbf{Bias-Corrected:}\\
\small{
\begin{verbatim}
target(X,`kitchen') :- cabinet2_wall2(X), not ab1(X).
target(X,`bedroom') :- bed2(X).
target(X,`bathroom') :- sink1_toilet1(X).
target(X,`bathroom') :- not bed1(X), not ab2(X).
target(X,`bedroom') :- bed1(X).
ab1(X) :- not cabinet1(X), bed2(X).
ab2(X) :- not wall1(X), not wall3(X).
\end{verbatim}
}
\subsection{\textit{dedrh} (`desert road', `driveway', `highway'):}
\medskip\noindent\textbf{Initial:}\\
\small{
\begin{verbatim}
target(X,`highway') :- not ground1(X), not ab4(X), not ab5(X).
target(X,`driveway') :- house3_building3(X), not ab7(X), not ab8(X).
target(X,`desert_road') :- not house2_sky2_trees2_building1_tree1_ground4_buildings1(X), not ab11(X), not ab12(X), not ab13(X).
target(X,`highway') :- road9(X), not ab14(X).
target(X,`driveway') :- house1(X).
target(X,`highway') :- road7(X).
ab1(X) :- not trees3(X), not road1(X).
ab2(X) :- not road15(X), trees3(X), not road13_car1(X), not sky1(X).
ab3(X) :- not house3_building3(X), not ab1(X), not ab2(X).
ab4(X) :- not road14(X), not ab3(X).
ab5(X) :- ground3(X), building2(X).
ab6(X) :- not road8(X),
    not road14(X), house2_sky2_trees2_building1_tree1_ground4_buildings1(X).
ab7(X) :- not house1(X), road5(X), not ab6(X).
ab8(X) :- not sky6_house4(X), sky5(X).
ab9(X) :- not trees1(X), sky4(X).
ab10(X) :- sky3(X), not ground2_road4(X), not house5_building4(X).
ab11(X) :- not mountain1(X), not road3(X), not ab9(X), not ab10(X).
ab12(X) :- road10(X), not road11(X).
ab13(X) :- road1(X), trees3(X), not road12(X).
ab14(X) :- not road6(X), road2(X).
\end{verbatim}
}

\medskip\noindent\textbf{Bias-Corrected:}\\
\small{
\begin{verbatim}
target(X,`highway') :- not ground1_road2(X), not house1(X), not sky2(X), not ab1(X).
target(X,`driveway') :- not sky2(X), not ab2(X), not ab3(X), not ab4(X).
target(X,`desert_road') :- sky2(X).
target(X,`desert_road') :- ground1_road2(X), not trees1(X), not ab5(X).
ab1(X) :- not road3(X), building1_house3_road5_car1(X).
ab2(X) :- not house1(X), not building1_house3_road5_car1(X), not house2_trees2(X).
ab3(X) :- road3(X), road1(X).
ab4(X) :- not house1(X), road4(X), not tree1_sky1(X).
ab5(X) :- not mountain1(X), grass1_trees3(X).

\end{verbatim}
}

\subsection{\textit{babe} (`bathroom', `bedroom'):}
\medskip\noindent\textbf{Initial:}\\
\small{
\begin{verbatim}
target(X,`bedroom') :- not mirror1(X,'1'), not bed4(X,'0').
target(X,`bathroom') :- not countertop1(X,'0').
target(X,`bedroom') :- not bed7(X,'0'), not ab1(X,'True').
target(X,`bathroom') :- bed3(X,'0').
target(X,`bedroom') :- bed3(X,'0').
target(X,`bedroom') :- not bathtub1(X,'1'), bed6(X,'1').
target(X,`bedroom') :- bed2(X,'1').
target(X,`bathroom') :- mirror1(X,'1').
ab1(X,'True') :- not bed1(X,'1'), not bathtub1(X,'0'), not bed5(X,'1').

\end{verbatim}
}

\medskip\noindent\textbf{Bias-Corrected:}\\
\small{
\begin{verbatim}
target(X,`bedroom') :- bed1(X).
target(X,`bathroom') :- not bed3(X), not ab1(X).
target(X,`bedroom') :- not sink1(X).
ab1(X) :- not sink1(X), bed2(X).
\end{verbatim}
}

\subsection{\textit{deh} (`desert road', `highway'):}
\medskip\noindent\textbf{Initial:}\\
\small{
\begin{verbatim}
target(X,`highway') :- sky2_trees1(X), not ab1(X), not ab2(X).
target(X,`desert_road') :- ground1(X), not ab3(X).
target(X,`highway') :- road3(X), not ab4(X).
target(X,`desert_road') :- sky1_road2(X), not ab5(X).
target(X,`highway') :- sky3(X).
target(X,`desert_road') :- road4(X).
target(X,`highway') :- road1(X).
target(X,`desert_road') :- not road7(X).
ab1(X) :- not road3(X), not trees2(X), sky4(X).
ab2(X) :- sky1_road2(X), ground1(X), not road5(X).
ab3(X) :- road9_trees3(X), road6(X), road13(X).
ab4(X) :- sky1_road2(X), not road10(X).
ab5(X) :- road11(X), road12(X), not road8(X).
\end{verbatim}
}

\medskip\noindent\textbf{Bias-Corrected:}\\
\small{
\begin{verbatim}
target(X,`highway') :- trees2(X).
target(X,`desert_road') :- ground1(X).
target(X,`desert_road') :- not road4(X).
target(X,`highway') :- not road1(X).
target(X,`highway') :- not road2(X).

\end{verbatim}
}
\end{document}